\documentclass{article}

\usepackage{pkg/spconf}
\usepackage{amsmath,graphicx}

\usepackage{hyperref}
\usepackage{color}
\usepackage{dirtytalk}  
\usepackage{glossaries}
\usepackage{subfigure}
\usepackage{float}
\usepackage{amssymb}
\usepackage{booktabs, multirow} 
\usepackage{soul}
\usepackage{changepage,threeparttable} 

\usepackage{tabularx}

\usepackage{xurl}

\title{Adapting the Hypersphere Loss Function from Anomaly Detection to Anomaly Segmentation}
\usepackage{tabularx}

\usepackage{xurl}
\name{Joao P. C. Bertoldo, Santiago Velasco-Forero, Jesus Angulo, Etienne Decencière}
\address{%
MinesParis, PSL University, Centre for mathematical morphology (CMM), 77300 Fontainebleau, France\\
\{jpcbertoldo, santiago.velasco, jesus.angulo, etienne.decenciere\}@minesparis.psl.eu
}

\glsdisablehyper

\newcommand{\params}{\omega}

\newcommand{\ascoremap}{A}
\newcommand{\ascoremapw}{\ascoremap^{\params}}
\newcommand{\gtmap}{Y}
\newcommand{\pushfunc}{p}
\newcommand{\Nimgs}{n}
\newcommand{\NpixelsPimg}{m}
\newcommand{\NpixelsLowres}{\NpixelsPimg'}
\newcommand{\img}{X}
\newcommand{\datapointsize}{d}
\newcommand{\featmapsize}{d_F}
\newcommand{\nn}{\phi_{\params}}
\newcommand{\hubert}{h}

\newcommand{\instance}{x}
\newcommand{\gt}{y}

\newcommand{\Reals}{\mathbb{R}}

\newglossaryentry{mvtecad}%
{
    name={MVTec-AD},
    first={MVTec Anomaly Detection Dataset (MVTec-AD)},
    description={}
}

\newglossaryentry{sgd}%
{
    name={SGD},
    first={Stochastic Gradient Descent (SGD)},
    description={}
}

\newglossaryentry{auroc}%
{
    name={AUROC},
    first={Area Under the Receiver Operator Characteristic curve (AUROC)},
    description={}
}

\newglossaryentry{ap}%
{
    name={AP},
    first={Average Precision (AP)},
    description={}
}

\newglossaryentry{hsc}%
{
    name={HSC},
    first={Hypersphere Classifier (HSC)},
    description={}
}

\newglossaryentry{fcdd}%
{
    name={FCDD},
    first={Fully Convolutional Data Description (FCDD)},
    description={}
}

\newglossaryentry{occ}%
{
    name={OCC},
    first={One-Class Classification (OCC)},
    description={}
}

\newglossaryentry{ad}%
{
    name={AD},
    first={Anomaly Detection (AD)},
    description={}
}

\newglossaryentry{sota}%
{
    name={SOTA},
    first={state-of-the-art (SOTA)},
    description={}
}

\newglossaryentry{svdd}%
{
    name={SVDD},
    first={Support Vector Data Description (SVDD)},
    description={}
}

\newglossaryentry{critdiff}%
{
    name={CD},
    first={Critical Difference (CD)},
    description={}
}

\begin{document}


\maketitle

\begin{abstract}
We propose an incremental improvement to Fully Convolutional Data Description (FCDD), an adaptation of the one-class classification approach from anomaly detection to image anomaly segmentation (a.k.a. anomaly localization). 
We analyze its original loss function and propose a substitute that better resembles its predecessor, the Hypersphere Classifier (HSC).   
Both are compared on the MVTec Anomaly Detection Dataset (MVTec-AD) -- training images are flawless objects/textures and the goal is to segment unseen defects -- showing that consistent improvement is achieved by better designing the pixel-wise supervision.
\end{abstract}


\section{Introduction}

\noindent
\gls{ad} aims to identify deviations or non-membership to a class (a semantic concept of normality) by learning from normal instances and none or very few examples of such irregularities.
In the past few years, this unsupervised problem has seen an increasing number of publications with novel approaches and several literature reviews \cite{ruff_unifying_2021,pang_deep_2021}. 

\noindent
Computer Vision in particular has received considerable attention thanks to public datasets like the \gls{mvtecad} \cite{bergmann_mvtec_2021}, which contains industrial images of objects/textures and defines an \gls{ad} problem both on the image level and on the pixel level. 
Image-wise \gls{ad} consists of flagging which images contain a defect, while pixel-wise \gls{ad}, a.k.a. \emph{anomaly segmentation} (or localization), consists of marking which pixels belong to a defect (ground truth masks are available).

\noindent
In this paper, we incrementally build on top of \gls{fcdd} \cite{liznerski_explainable_2021}, a model that outputs, by design, a heatmap (see Figure~\ref{fig:fcdd-architecture}) from the last feature map of a convolutional neural network by applying a \gls{svdd}-inspired loss pixel-wise.
We reformulate the \gls{hsc} loss function to the segmentation task (initially designed for the image-wise setting), and \gls{critdiff} diagrams using the Wilcoxon signed rank test (a non-parametric version of the paired T-test) demonstrate consistent improvement in performance with less training epochs.





\section{Related work}

\noindent
\gls{fcdd} \cite{liznerski_explainable_2021} is inspired by modifications of \gls{svdd}, a \gls{occ} model. 
In this section we summarize the predecessors that led to its formulation, focusing on their incremental changes. 
For the sake of clarity, some details have been omitted, such as regularization terms.
An extensive review of \gls{occ} models can be found in \cite{perera_one-class_2021}.


\noindent
\gls{occ} was first formulated as an adaptation of \emph{Support Vector Machines} (SVMs) that learns  to separate anomalous data points from the origin with a hyperplane \cite{scholkopf_estimating_2001} or a hypersphere \cite{tax_support_2004} in the feature space.
The latter, \emph{\gls{svdd}} \cite{tax_support_2004}, minimizes the radius of a hypersphere by incentivizing (via slack variables) the normal/anomalous data points to fall inside/outside the hypersphere.
In other words, normal and anomalous instances are penalized in opposite directions, therefore (respectively) pulling towards and pushing away from the center -- notice that this formulation assumes supervision.



\noindent
One-Class Deep SVDD \cite{ruff_deep_2018}, a deep-learning model inspired by \gls{svdd}, applies a neural network $\nn$ with parameters $\omega$ to normal instances $\instance_i \in \Reals^{\datapointsize}$, obtaining feature vectors $\nn(\instance_i) \in \Reals^{\featmapsize}$, which are trained by minimizing

\begin{equation}\label{eq:deepsvdd}
    \frac{1}{n} \sum_{i=1}^{n} \lVert \nn(\instance_i) - a \rVert^2
    \quad ,
\end{equation}

\noindent
where $a\in \Reals^{\featmapsize}$ is the center of the hypersphere in the feature space and $n$ is the number of instances in the batch.
In other words, the mean squared Euclidean distance to the center $a$ is minimized, and this distance is used as an anomaly score at inference time.
Notice that this version does \textit{not} consider anomalous instances, which makes the training less stable and (eventually) leads to feature collapse ($\nn$ converging to a constant function).




\noindent
Among several strategies proposed to prevent feature collapse, \cite{ruff_deep_2020,ruff_rethinking_2021} proposed an extension of \eqref{eq:deepsvdd} that brings in back the possibility of using supervision (therefore using normal \textit{and} anomalous instances at training time) by applying 

\begin{equation}\label{eq:pushfunc}
\pushfunc(\cdot) = - \log ( 1 - \exp ( - \cdot  ) ) \; ,
\end{equation}

\noindent
to the anomalous instances' scores, thus creating an incentive that \textit{pushes} their (learned) feature vectors away from the center (we call $\pushfunc: \Reals^{+} \to \Reals^{+}$ the \say{push} function).
Besides the squared distance $\lVert \cdot \rVert^2$ in Equation~\ref{eq:deepsvdd} was replaced by the Pseudo-Hubert function $\hubert ( \cdot ) = \sqrt{ \lVert \cdot \rVert^2 + 1} - 1 \in \Reals^{+}$ as it is more robust to outliers, resulting in the \glsfirst{hsc} loss function: 

\begin{equation}\label{eq:hsc}
    \frac{1}{n} 
    \sum_{i=1}^{n} 
    (1 - \gt_i) \, 
    \hubert
    \left( \nn(\instance_i) - a \right) 
    + 
    \gt_i \,
    \pushfunc
    \left(
        \hubert
        \left( \nn(\instance_i) - a \right)
    \right)
    \,,
\end{equation}

\noindent 
where $\gt_i$ is $0$/$1$ if $\instance_i$ is normal/anomalous respectively. 





\begin{figure}[t]
\begin{minipage}[b]{1.0\linewidth}
\centering
\centerline{\includegraphics[width=\linewidth]{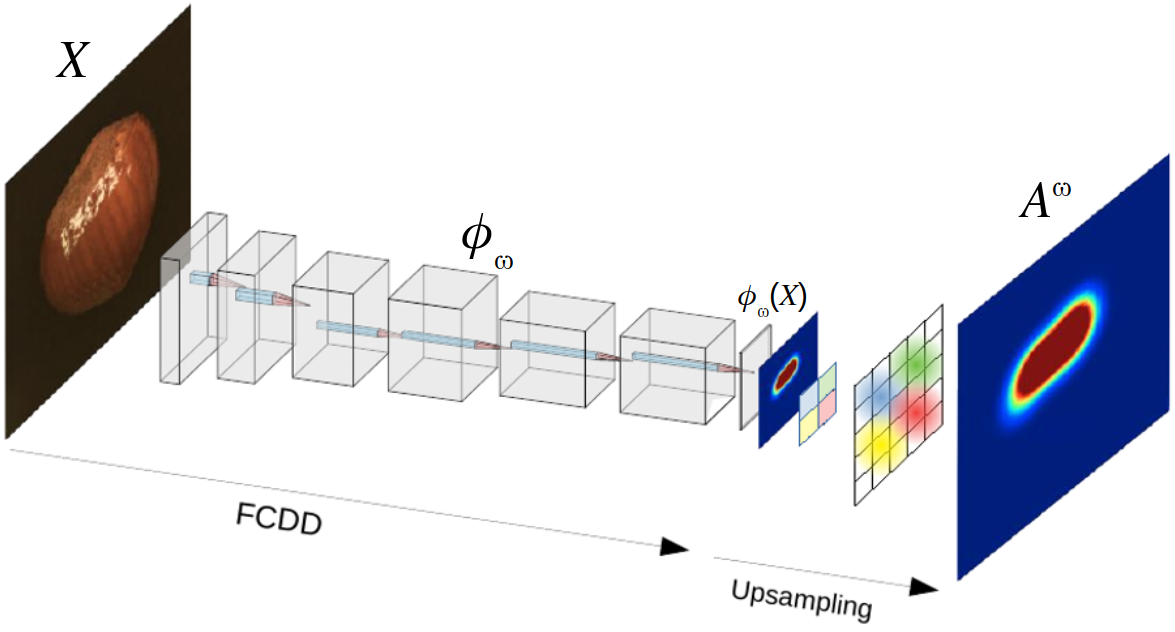}}
\end{minipage}
\caption{
Fully Convolutional Data Descriptor (FCDD). A pre-trained (and fine-tuned) convolutional neural network without fully connected layers is combined with an extra 1x1 convolution layer to generate locality-preserving feature vectors. 
In the last (low resolution) layer, the distances to the hypersphere's center are upsampled to the network's input resolution and used as an anomaly score heatmap.
Adapted from \cite{liznerski_explainable_2021}.
}
\label{fig:fcdd-architecture}
\end{figure}

\noindent
Finally, \gls{fcdd}, illustrated in Figure~\ref{fig:fcdd-architecture}, reused the \gls{hsc} to directly segment anomalies by design.
The first layers of a backbone convolutional neural network (e.g. VGG) feed a feature map to an extra (learnable) 1x1 convolution added on top of it.
At inference time, a low-resolution anomaly score map, or \say{heatmap}, is generated (from the pixel-wise vectors' distances to the center of the hypersphere) and upsampled to the input resolution -- a more formal formulation is presented in the next section. 

\noindent
Numerous other approaches (unlike the \gls{occ}) exist and will not be covered here, but a few major research branches are worth noting. 
Autoencoders (reconstruction models more generally) have been extensively studied and improved for anomaly detection tasks, and some variations still remain in the \gls{sota} \cite{zavrtanik_draem_2021}.
Combinations of deep learning-based feature extraction with classic machine learning models like k-Nearest Neighbors \cite{cohen_sub-image_2020,roth_towards_2022} and multivariate Gaussian \cite{rippel_gaussian_2021,defard_padim_2021} also achieve effective results despite their simplicity, and recent improved versions are also in the \gls{sota} \cite{jang_n-pad_2022,bae_image_2022}.
However, nowadays most \gls{sota} methods are based on pre-trained feature extractors combined with normalizing flows \cite{kim_altub_2022}.

\noindent
Our work aims to contribute towards closing the performance gap between \gls{occ} models and the current \gls{sota} by finding a better loss function for \cite{liznerski_explainable_2021}, explained in Section~\ref{sec:methods}, and our results, in Section~\ref{sec:results}, demonstrate its effectiveness.

\section{Loss function}\label{sec:methods}

\noindent
In this section we first present \gls{fcdd}'s loss function \eqref{eq:fcdd-loss-push}, then propose our version \eqref{eq:fcdd-loss-per-batch}, and finally describe its rationale by comparing them.

\noindent
Let $\img \in \Reals^{\Nimgs \times \NpixelsPimg}$ be a batch of $\Nimgs$ images of $\NpixelsPimg$ pixels (spatial dimensions confounded) and $\gtmap \in \{0, 1\}^{\Nimgs \times \NpixelsPimg}$ its respective batch of ground truth anomaly masks ($0$ means \say{normal}, $1$ means \say{anomalous}), where $\img_{ij}$ refers to the pixel indexed by $j \in \{ 1, ..., \NpixelsPimg \}$ in the image $\img_{i}$, indexed by $i \in \{ 1, ..., \Nimgs \}$ ($\gtmap_{ij}$ defined accordingly).
 
\noindent
First, the neural network $\nn: \Reals^{\NpixelsPimg} \to \Reals^{\featmapsize \times \NpixelsLowres}$, with parameters $\omega$, generates a feature map $\nn\left(\img_i\right) \in \Reals^{\featmapsize \times \NpixelsLowres}$ with low-resolution $\NpixelsLowres$ ($\NpixelsLowres < \NpixelsPimg$) and $\featmapsize$ features for each pixel.
Then, a $1 \times 1$ convolution with bias outputs a single value per pixel -- the following notation considers this convolution (and its parameters) included in $\phi$, and the center $a$ has been omitted as it is implicitly embedded in it.
Finally, the Gaussian kernel-based upsampling operator outputs the anomaly score map $\ascoremapw_i = \texttt{upsample}(\hubert(\nn(\img_i))) \in (\Reals^+)^{\NpixelsPimg}$ (indices defined as for $\img_{ij}$), where $\hubert(\cdot)$ is the Pseudo-Hubert function applied pixel-wise. 
\cite{liznerski_explainable_2021} optimizes the parameters $\omega$ by minimizing the loss function  

\begin{equation}\label{eq:fcdd-loss-push}
    \frac{1}{n} 
    \sum_{i=1}^{n}
    \left[
        \left(
            \frac{1}{\NpixelsPimg} 
            \sum_{j=1}^{\NpixelsPimg} 
            (1 - \gtmap_{ij}) 
            \ascoremapw_{ij}
        \right)
        + \pushfunc
        \left(
            \frac{1}{\NpixelsPimg} 
            \sum_{j=1}^{\NpixelsPimg}
            \gtmap_{ij} 
            \ascoremapw_{ij} 
        \right)
    \right]
    .
\end{equation}
 


\newcommand{\pixelidxset}{J}
\newcommand{\pixelidxsetNorm}{\pixelidxset^{0}}
\newcommand{\pixelidxsetAnom}{\pixelidxset^{1}}
\newcommand{\pixelidxinrange}{\{ 1, ..., \NpixelsPimg \}}
\newcommand{\imgidxinrange}{\{ 1, ..., \Nimgs \}}
\newcommand{\balancingfactor}{\frac{|\pixelidxsetNorm|}{|\pixelidxsetAnom|}}

\newcommand{\sumOfN}{\frac{1}{\Nimgs}\sum_{i=1}^{\Nimgs}}
\newcommand{\sumOfM}{\frac{1}{\NpixelsPimg}\sum_{j=1}^{\NpixelsPimg}}
\newcommand{\sumOfNM}{\frac{1}{\Nimgs\NpixelsPimg}\sum_{i=1}^{\Nimgs}\sum_{j=1}^{\NpixelsPimg}}

We propose, as a substitute to Equation \eqref{eq:fcdd-loss-push}, the following loss function

\begin{equation}\label{eq:fcdd-loss-per-batch}
    \frac{1}{n \NpixelsPimg} 
    \sum_{i=1}^{n}
    \sum_{j=1}^{\NpixelsPimg}  \;
        (1 - \gtmap_{ij}) \;
        \ascoremapw_{ij}
        + 
        \balancingfactor \;
        \gtmap_{ij} \;
        \pushfunc \left( \ascoremapw_{ij} \right)
    \; ,
\end{equation}

\noindent
where $\pixelidxset^{y} = \{ \, (i, j) \in \pixelidxinrange \times \imgidxinrange \, | \, \gtmap_{ij} = y \, \}$ for $y \in \{ \, 0, 1 \,\}$ are, respectively, the sets of image and pixel indices of all normal and anomalous pixels in a batch.

\noindent
Equation \eqref{eq:fcdd-loss-push} is the cross-image average ($\sumOfN$) of the sum of two terms that \textit{look} like per-image averages ($\sumOfM{}$) of anomaly scores on the normal pixels and on the anomalous pixels (left and right respectively).
In fact, $\gtmap_{ij}$ zeroes part of the iterands, so the $\sumOfM{}$ terms do not converge to any expected value. 
By correcting that, one would have $\pushfunc{}$ applied to the \textit{average score} of the anomalous pixels (not each pixel individually), which interferes with the backpropagation: (very) high-scored pixels can compensate the others blocking the propagation through (very) low-scored pixels.
Therefore we propose to apply $\pushfunc$ on each pixel individually.

\noindent
With these modifications, one would achieve a nested average (cross-image outside, cross-pixel inside).
However, we propose to take the average directly on all the pixels in the batch ($\sumOfNM$) because otherwise the nested averages would give more weight to smaller anomalies.
Finally, we scale the anomalous pixels' scores by a factor $\balancingfactor$, compensating the imbalance of normal/anomalous pixels, which our tests have shown to be effective (details omitted for brevity) -- other (more modern and effective) techniques to compensate imbalanced datasets exist but we deliberately do not focus on that subject.

\section{Experiments}

\begin{figure}[t]
\begin{minipage}[b]{1.0\linewidth}
\centering
\centerline{\includegraphics[width=\linewidth]{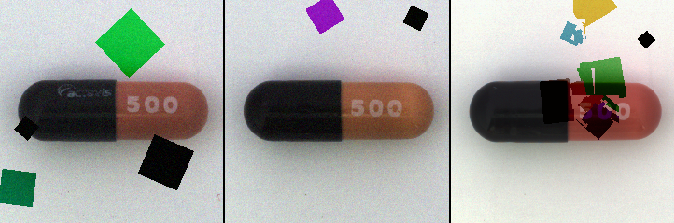}}
\end{minipage}
\caption{Synthetic anomalies generated with confetti noise.}
\label{fig:confetti}
\end{figure}

\noindent
We aim to demonstrate that our proposed loss function improves from its baseline, therefore we test it with the same procedure as in \cite{liznerski_explainable_2021} (a few differences specified below) on \gls{mvtecad} -- 10 categories of object (e.g. \say{transistor}) and 5 categories of texture (e.g. \say{leather}), each category being trained and tested independently.

\noindent
Procedure: a VGG-11, pre-trained on ImageNet, frozen up to the third max-pooling and cut before the forth max-pooling is used; the network is fine-tuned using \gls{sgd} with a batch size of 128 images, Nesterov momentum of 0.9, weight decay of 10$^{-4}$, and initial learning rate of 10$^{-3}$ decreasing by a factor of 1.5\% at each epoch.

\noindent
One epoch is defined as 10 iterations over the training set with each image having 50\% chance to be replaced by an anomalous image (before the data preprocessing/augmentation described below), which can be synthetic or real depending on the experiment (never both in the same training).
Synthetic anomalies (unsupervised setting): automatically generated by superposing a confetti noise -- randomly sized, placed, and colored squares -- over the normal image, and the modified pixels are considered anomalous (Figure~\ref{fig:confetti}).
Real anomalies (semi-supervised): one (randomly chosen) image of each anomaly type (between 1 and 6 types according to the \gls{mvtecad} category) is transferred from the test set to the training set.
Notice that our focus is on the unsupervised setting but the semi-supervised setting (\say{a few} anomalies available) is also tested for reference.

\noindent
Each experiment was repeated six times with different seeds, and the performance is measured in terms of run-wise average of two metrics: pixel-wise \gls{auroc} and pixel-wise \gls{ap}.
Differences with \cite{liznerski_explainable_2021}: we run the gradient descent for 50 epochs instead of 200, and the synthetic anomalies are not smoothed on the borders.


\noindent
The training images are pre-processed and augmented in the following sequence: resize to 240 x 240; with 50\% chance each, resize to 224 x 224 or crop to 224 x 224 in a random position; apply color jitter with all parameters set to 0.04 or all parameters set to 0.0005 with 50\% chance each; apply Gaussian noise with zero mean and standard deviation set to 10\% of the image's pixel values' standard deviation (all channels confounded) with 50\% chance on each pixel; apply local contrast normalization (all channels confounded); normalize the data between 0 and 1 (min-max normalization).
The test images are resized to 224 x 224 then the last two last two steps from above are applied.

\noindent
We ran our experiments on two cluster nodes PowerEdge C4130 (processor Xeon E5-2680 V4, 7 cores, 2.4GHz), each with four GPUs Tesla P100-SXM2 (16 GiB), and we used Weights \& Biases to monitor our experiments.
The minimum GPU memory requirement is 5.6 GiB, and the (category-specific) average run time spans from 33 minutes to 2 hours.
We used the code publicly provided by \cite{liznerski_explainable_2021} with modifications to integrate the losses described in Section~\ref{sec:methods}.

\section{Results}\label{sec:results}

\begin{figure*}[tbh]
\begin{minipage}[t]{\textwidth}
\centering
    \begin{minipage}[t]{.49\textwidth}
    \centering
    \centerline{\includegraphics[width=\linewidth]{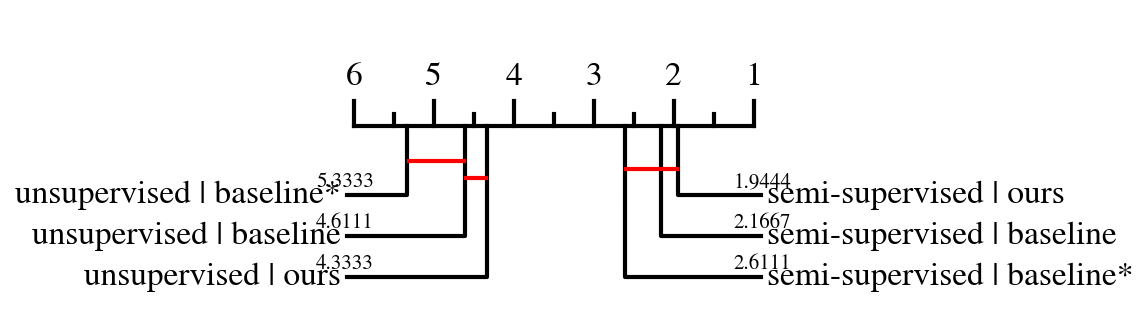}}
    \centerline{(a) Pixel-wise AUROC}\medskip
    \end{minipage}
    \hfill
    \begin{minipage}[t]{.49\textwidth}
    \centering
    \centerline{\includegraphics[width=\linewidth]{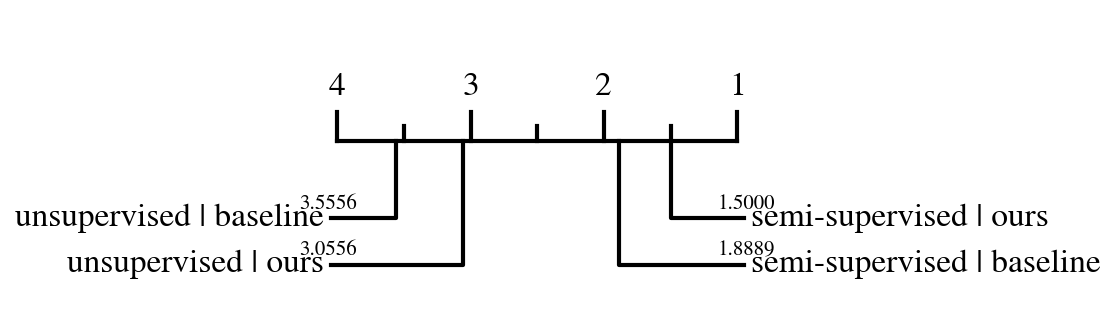}}
    \centerline{(b) Pixel-wise Average Precision}\medskip
    \end{minipage}
\caption{
Critical difference diagrams from the category-specific scores on the MVTec AD dataset using the Wilcoxon signed-rank test with the Bonferroni-Holm correction with a confidence level of $\alpha =10\%$ at for each pair of methods. 
Values on the scale are average rankings, and the methods within a same horizontal bar (a clique) are \textit{not}  significantly different according to the pair-wise statistical test.
The scores of the methods marked with ``*" are from \cite{liznerski_explainable_2021}.
}
\label{fig:03-04-cd-diagrams}
\end{minipage}
\end{figure*}

\noindent
Tables \ref{tab:02bis-perf-short-auroc} and \ref{tab:03bis-perf-short-avg-precision} compare the cross-category average results obtained with the baseline loss from \cite{liznerski_explainable_2021} (Equation~\ref{eq:fcdd-loss-push}) and ours (Equation~\ref{eq:fcdd-loss-per-batch}) in terms of, respectively, \gls{auroc} and \gls{ap}.
The baseline results in terms of \gls{auroc} are both copied from \cite{liznerski_explainable_2021} (with ``*") and reproduced by us.
Category-specific scores available on GitHub: \href{https://gist.github.com/jpcbertoldo/8db759487279a5eec6fd0fc36764a4b4}{\nolinkurl{gist.github.com/jpcbertoldo/8db759487279a5eec6fd0fc36764a4b4}}.

\noindent
Figure \ref{fig:03-04-cd-diagrams} shows \gls{critdiff} diagrams (with each metric) using the pairwise Wilcoxon signed-rank test with Bonferroni-Holm correction with a (nominal) significance level of $\alpha = 10\%$ (probability of a type I error) considering the two-sided alternative hypothesis.

\noindent
The Wilcoxon signed-rank test's null hypothesis is that two methods are expected to rank similarly (\say{they have the same performance}) for any dataset, and non-parametrically compares their cross-dataset performance differences -- notice that the categories in \gls{mvtecad} can be considered as independent datasets because they are trained and tested individually.
In other words, it provides a stronger statistical argument (than comparing average performances) because rejecting its null hypothesis (i.e. methods that are not connected by a horizontal red bar in Figure~\ref{fig:03-04-cd-diagrams}) means that a method is (in general) significantly better than another regardless of the dataset (all other factors fixed).






\begin{threeparttable}[tb]
\centering
\caption{%
Pixel-wise Area Under the ROC curve (AUROC) scores (values in percentage) on the MVTec AD dataset.
Cross-category average scores and cross-category average standard deviation (in parenthesis).
Under the same supervision (group of three columns), the best performance is in bold.
The columns "baseline*" is from \cite{liznerski_explainable_2021}.
}
\label{tab:02bis-perf-short-auroc}
\begin{tabularx}{\linewidth}{XXXX}
 \multicolumn{4}{c}{semi-supervised} \\
  & baseline* & baseline & ours \\
 objects & 95.2* & 95.7 (1.4) & \bfseries 95.8 (1.4)  \\
 textures & 97.0* & 97.8 (0.4) & \bfseries 97.9 (0.4) \\
 all & 95.8* & 96.4 (1.0) & \bfseries 96.6 (1.0) \\
\end{tabularx}%
\vspace{2mm}%
\begin{tabularx}{\linewidth}{XXXX}
 \multicolumn{4}{c}{unsupervised} \\
  & baseline* & baseline & ours \\
 objects & 91.0* & \bfseries 92.5 (1.1) & 92.5 (1.0) \\
 textures & 92.8* & 92.8 (1.3) & \bfseries 94.1 (0.7) \\
 all & 91.6* & 92.6 (1.1) & \bfseries 93.1 (0.9) \\
\end{tabularx}
\end{threeparttable}




\section{Discussion}

\noindent
As shown in Tables \ref{tab:02bis-perf-short-auroc} and \ref{tab:03bis-perf-short-avg-precision} the our modifications provided, on average, an increase in performance -- categories \say{metal nut} and \say{tile} had the highest improvements (see the link to the full per-category table in Section~\ref{sec:results}).
The average scores over object, texture, and all classes confounded show improvement on both metrics and settings, while their average variance decreased in the unsupervised setting (cf. standard deviations in parenthesis).

\noindent
Figure~\ref{fig:03-04-cd-diagrams} confirms that indeed our loss function is ranked better than the previous one, and, in particular, Figure~\ref{fig:03-04-cd-diagrams}.b shows that, in terms of \gls{ap}, ours is significantly better across different datasets.

\noindent
Finally, we also observe that semi-supervised setting almost always performs better than the unsupervised one even having very few anomalous images (1 up to 7 seven depending on the class), so the effect of using proper anomalies in the training set is a major factor for performance improvement.





\begin{threeparttable}[tb]
\centering
\caption{
Pixel-wise Average Precision (AP) scores (values in percentage) on the MVTec AD dataset.
Cross-category average scores and cross-category average standard deviation (in parenthesis).
Under the same supervision (group of two columns), the best performance is in bold.
}
\label{tab:03bis-perf-short-avg-precision}
\begin{tabularx}{\linewidth}{Xcc|cc}
  & \multicolumn{2}{c}{semi-supervised} & \multicolumn{2}{c}{unsupervised} \\ 
  & baseline & ours & baseline & ours \\
 objects & 52.0 (5.8) & \bfseries 53.4 (5.8) & 36.6 (4.3) & \bfseries 38.7 (2.3) \\
 textures & 50.3 (4.5) & \bfseries 54.7 (4.1) & 37.8 (2.8) & \bfseries 40.2 (2.0) \\
 all & 51.4 (5.3) & \bfseries 53.8 (5.2) & 37.0 (3.8) & \bfseries 39.2 (2.2) \\
\end{tabularx}
\end{threeparttable}




\section{Conclusion}




\noindent
This paper introduced an adaptation for the hypersphere loss functions for anomaly segmentation on images. We have highlighted the drawbacks of the loss function originally used in \gls{fcdd} \cite{liznerski_explainable_2021} (Equation~\ref{eq:fcdd-loss-push}) and proposed an alternative  loss function (Equation~\ref{eq:fcdd-loss-per-batch}). 

\noindent
An evaluation of the statistical significance of the proposition is carried out using pairwise Wilcoxon signed-rank tests, summarized in \gls{critdiff} diagrams (Figure~\ref{fig:03-04-cd-diagrams}). Our proposition reaches, on average, better performances (Tables~\ref{tab:02bis-perf-short-auroc} and~\ref{tab:03bis-perf-short-avg-precision}) and is, in general, better ranked regardless of the image category in \gls{mvtecad}. 

\noindent
We highlight that the new loss introduced in this paper can be used with other neural network architectures better designed for segmentation (e.g. U-Net and its improved versions), which we expect to provide further higher improvements.

\vfill
\pagebreak
\bibliographystyle{pkg/IEEEbib}
\bibliography{refsselected}  

\end{document}